\documentclass[letterpaper, 10pt, conference]{ieeeconf}  

\IEEEoverridecommandlockouts                              

\usepackage[T1]{fontenc}
\usepackage[utf8]{inputenc}
\usepackage{amsthm}
\usepackage{times}
\usepackage{multicol}
\usepackage[bookmarks=true]{hyperref}
\usepackage{xcolor}
\usepackage{hyperref}
\usepackage{amsmath, amssymb}
\usepackage{amsfonts}
\usepackage{graphicx}
\usepackage{siunitx}
\usepackage{standalone}
\usepackage{booktabs}
\usepackage[ruled,vlined,linesnumbered,noend]{algorithm2e}
\usepackage{mdframed}
\usepackage{fancyvrb,multirow}
\usepackage{soul}
\usepackage{dsfont,mathabx}
\usepackage{array, booktabs}
\usepackage{subfigure}
\usepackage{booktabs}
\usepackage{makecell}
\usepackage{threeparttable}

\theoremstyle{definition}
\newtheorem{assumption}{Assumption}

\newtheorem{remark}{Remark}

\newtheorem{problem}{Problem}

\title{\LARGE \bf Uncertainty-bounded Active Monitoring of Unknown\\ Dynamic Targets in Road-networks with Minimum Fleet}

\author{Shuaikang Wang$^1$, Yiannis Kantaros$^2$ and Meng Guo$^1$
  \thanks{
    The authors are with $^1$the College of Engineering,
    Peking University, Beijing 100871, China;
    and $^2$the Department of Electrical and Systems Engineering,
    Washington University at St. Louis, St. Louis, MO, 63130, USA.
    This work was supported by $^1$the National Natural Science Foundation
    of China (NSFC) under grants 62203017, U2241214, T2121002;
    $^1$the Fundamental Research Funds for the central universities;
    and $^2$the ARL grant DCIST CRA W911NF-17-2-0181.
    Contact: {\tt\small meng.guo@pku.edu.cn}.}
}

\begin{document}
\maketitle
\thispagestyle{empty}
\pagestyle{empty}


\begin{abstract}
  Fleets of unmanned robots can be beneficial for the long-term monitoring
  of large areas, e.g., to monitor wild flocks, detect intruders, search and rescue.
  Monitoring numerous dynamic targets in a collaborative and efficient way is a challenging problem
  that requires online coordination and information fusion.
  The majority of existing works either assume a passive all-to-all
  observation model to minimize the summed uncertainties
  over all targets by all robots,
  or optimize over the jointed discrete actions while neglecting the dynamic
  constraints of the robots and unknown behaviors of the targets.
  This work proposes an online task and motion coordination algorithm that ensures an explicitly-bounded
  estimation uncertainty for the target states, while minimizing the average number of active robots.
  The robots have a limited-range perception to actively track a limited number of
  targets simultaneously, of which their future control decisions are all unknown.
  It includes: (i) the assignment of monitoring tasks,
  modeled as a flexible size multiple vehicle routing problem with time windows (m-MVRPTW),
  given the predicted target trajectories with uncertainty measure in the road-networks;
  (ii) the nonlinear model predictive control (NMPC) for optimizing the robot trajectories
  under uncertainty and safety constraints.
  It is shown that the robots can switch between active and inactive roles dynamically online
  as required by the unknown monitoring task.
  The proposed methods are validated via large-scale simulations
  of up to $100$ robots and targets.
\end{abstract}

\section{Introduction}\label{sec:intro}
With the rapid development of robotic perception and motion techniques,
mobile robots have been deployed to monitor areas that are otherwise too large or hostile for humans.
Fleets of unmanned aerial vehicles (UAVs) and unmanned ground vehicles (UGVs)
can be particularly suitable for this purpose, via wireless communication and collaborative execution,
to e.g., map time-varying fields~\cite{khodayi2019model,khodayi2019distributed,guo2018multirobot},
track moving targets~\cite{tzes2023graph,kalluraya2023multi} and explore unknown territory~\cite{burgard2005coordinated}.
As a fundamental technique to many applications,
active monitoring of unknown dynamic targets has attracted significant attention,
see~\cite{atanasov2014information,dames2012decentralized,chung2006decentralized,schlotfeldt2018anytime}.
It remains a challenging problem as it involves the online coordination of monitoring tasks among the robots,
and the real-time adaptation of their trajectories w.r.t. the observed behaviors of the targets.

\subsection{Related Work}\label{subsec:intro-related}
As the most relevant work,~\cite{atanasov2014information, le2009trajectory} pioneered
the problem of active information acquisition (AIA) with sensing robots.
The proposed solutions exploit the separation principles under the linear observation model
and utilize an offline search algorithm over the joint state-and-information space,
via the forward value iteration (FVI) and later the reduced forward value iteration (RVI).
This line of methods {has} been extended to multi-robot fleets,
where an optimal, non-myopic, and centralized sampling-based solution is proposed
in our earlier work~\cite{kalluraya2023multi, kantaros2019asymptotically};
and decentralized myopic planners can be found
in~\cite{dames2012decentralized,chung2006decentralized,schlotfeldt2018anytime, chen2024accelerated}.
The target tracking problem is combined with robot localization
in~\cite{sung2020distributed} for only one target.
Recent work~\cite{tzes2023graph} utilizes graph neural networks to
learn distributed and scalable action policies from the optimal planner.
However, most of these works assume a \emph{fixed} size of the robotic team
with discretized control inputs and a linear observation model,
and more importantly the targets move with a known model,
i.e., known dynamics and control inputs.
Instead, behaviors of the considered targets including velocities and paths
within the road network are all unknown,
thus {requiring} online prediction and tracking.

\begin{figure}[t]
  \centering
  \includegraphics[width=0.98\linewidth]{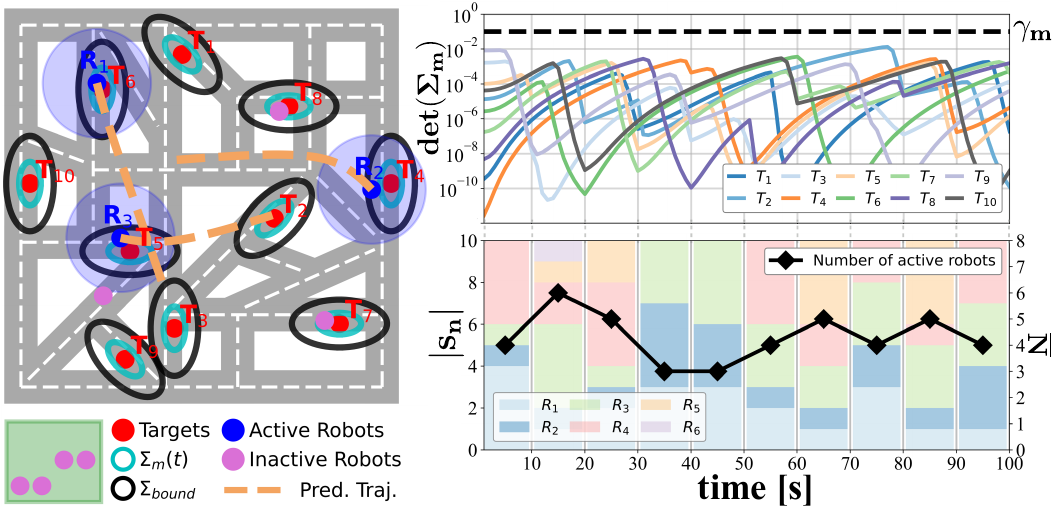}
  \vspace{-3mm}
  \caption{Illustration of the considered scenario.
    \textbf{Left}: $3$ UAVs (in blue) are actively
    monitoring~$10$ targets (in red) within a road network,
    with $7$ inactive ones (in orchid) that are recruited later online;
    \textbf{Top-right}: Evolution of estimation uncertainty for all targets
    below a specified upper-bound (in black);
    \textbf{Bottom-right}: Number of active UAVs in the fleet
    and number of targets assigned to each active UAV.}
  \label{fig:overall}
  \vspace{-5mm}
\end{figure}

On the other hand, another line of research formulates the
multi-target monitoring problem as a sequential high-level assignment problem,
where robots are assigned to targets for reducing uncertainty.
A distributed Hungarian method is proposed in~\cite{chopra2017distributed},
which however only applies to one-to-one assignments.
Numerous works~\cite{sung2020distributed, tokekar2014multi, zhou2019sensor}
reformulate it as a simultaneous action and target assignment (SATA) problem,
and propose distributed approximation algorithms such as the linear programs.
However, the target trajectories are often assumed to be given,
and the robots move synchronously with a set of pre-defined motion primitives.
The work~\cite{dames2017detecting} considers a complementary problem to this work,
which maximizes the number of targets being tracked by a fixed number of robots
via a 2-approximation greedy solution.
Recent work~\cite{zhou2018resilient, zhou2023robust} studies
the robust assignment strategies
against possible attacks in communication and sensing,
which also assumes a synchronized team motion with a set of finite primitives.
Instead, asynchronous and timed robot trajectories are computed recursively online
in this work given the observed target behaviors.

\subsection{Our Method}\label{subsec:intro-our}
This work addresses the dynamic monitoring problem from a practical perspective,
i.e., \emph{how many robots are needed to monitor a set of dynamic targets with an
upper-bounded uncertainty, and how to control their trajectories}.
Different from many existing {works}, these targets follow a constant-velocity model,
but with unknown velocity on each road, and unknown path within the road network.
The proposed solution consists of two-layers:
(i) the first layer explicitly formulates the assignment problem of the
active monitoring tasks, under the constraints that each robot can only monitor
a maximum number of targets simultaneously.
The assignment scheme first over-approximates the monitoring quality
given the predicted trajectories and uncertainties of all targets,
then formulates a flexible-size multiple vehicle routing problem with time windows (m-MVRPTW);
(ii) the second layer then formulates a constrained optimal control problem
given the assignment results.
The robot trajectories are optimized online via the distributed
nonlinear model predictive control (NMPC),
where the dynamical constraints and predicted target trajectories are
all incorporated.
Both optimizations can be solved via off-the-shelf solvers.

Main contribution of this work is three-fold:
(i) the novel formulation of active monitoring problem for unknown targets
with an upper-bounded uncertainty and a minimum fleet.
To the best of our knowledge, this problem has not been addressed in related work;
(ii) the proposed two-level solution adapts the fleet size and the robot trajectories
online, according to real-time measurements of the target behavior;
(iii) the scalable solution that allows the deployment of a few UAVs to monitor a large number of dynamic targets.

\section{Problem Description}\label{sec:problem}

\subsection{Robots and Targets}\label{subsec:robots}
Consider a team of~$N$ autonomous robots that coexist within a complex
workspace~$\mathcal{W}\subset \mathbb{R}^{3}$, where the total number~$N$ might be time-varying thus $N(t)$ for time~$t>0$.
Without loss of generality,
each robot~$n \in \mathcal{N}(t)\triangleq \{1,\cdots,N\}$ follows the same generic nonlinear dynamics:
\begin{equation}\label{eq:robot-model}
  \mathbf{x}_{n}(t+1) \triangleq f\big(\mathbf{x}_{n}(t),\, \mathbf{u}_{n}(t)\big),
\end{equation}
where~$f(\cdot)$ is the dynamic model;
$\mathbf{x}_{n}(t)\in \mathcal{W}$ and $\mathbf{u}_{n}(t)\in \mathcal{U}$
are the state and control input of robot~$i$ at time step~$t\geq 0$.
It is assumed in this work that the robots can communicate freely via a distributed network.
Note that the total number~$N$ is not fixed, rather a part of the objective to minimize.

On the other hand, there are~$M$ targets that move dynamically within a
known road network.
Denote by~$\mathcal{R}\triangleq (\mathcal{V},\, \mathcal{E})$ this network,
where~$\mathcal{V}\subset \mathcal{W}$ is the set of hub points and
$\mathcal{E}\subset \mathcal{V}\times \mathcal{V}$ is a set of \emph{straight} roads connecting these hubs.
Each target follows a sequence of roads to move from its starting point to the destination points,
which are all \emph{unknown} to the robots.
On different roads,
each target~$m \in \mathcal{M} \triangleq \{1,\cdots,M\}$ follows the
same constant-velocity model~\cite{atanasov2014information}
but with different velocities, i.e.,
\begin{equation}\label{eq:target-model}
  \mathbf{y}_{m}(t+1) \triangleq G\,\mathbf{y}_{m}(t)  + \mathbf{w}_m(t),
\end{equation}
where~$G\triangleq [\mathbf{I}_2,\,h\mathbf{I}_2;\mathbf{0},\,\mathbf{I}_2]$
with sampling time~$h>0$;
$\mathbf{y}_{m}(t)\in \mathcal{W}$ is the target state at~$t\geq 0$,
and $\mathbf{w}_{m}(t)\sim \mathcal{N}(0,\,W_m)$ is the Gaussian noise with covariance~$W_m$.
Namely, \emph{the target may switch to a different road and move with an unknown velocity,
  i.e., following an \textbf{unknown} path.}

\subsection{Active Sensing Model}\label{subsec:obsrv-model}
Moreover, each robot~$n\in \mathcal{N}$ is equipped with a perception module
to \emph{actively track} the targets via action~$\mathbf{s}_n:\mathbb{Z}\rightarrow 2^{\mathcal{M}}$:
\begin{equation}\label{eq:observation-model}
  \mathbf{z}_{nm}(t) \triangleq h\big(\mathbf{x}_{n}(t), \, \mathbf{y}_{m}(t)\big)
  + \mathbf{v}(t),\, \forall m \in \mathbf{s}_n(t);
\end{equation}
where~$h(\cdot)$ maps the current states of robot~$n$ and target~$m$
to the measurement~$\mathbf{z}_{nm}$,
i.e., robot~$n$ w.r.t. target~$m$;
and $\mathbf{v}(t)\sim \mathcal{G}(0,\,V\big(\mathbf{x}_{n}(t), \, \mathbf{y}_{m}(t)\big))$ is the Gaussian noise with a covariance~$V(t)$
that varies by the robot and target states.

\begin{remark}\label{remark:nonline-observe}
  Note that the observation model $h(\cdot)$ is a general nonlinear and differentiable function,
  not necessarily linear w.r.t.~$\mathbf{y}_m$.
  Thus, the separation principle does not hold anymore~\cite{atanasov2014information}.
  Consequently, the offline approaches in~\cite{dames2012decentralized,
    schlotfeldt2018anytime,kantaros2019asymptotically} {cannot} be applied;
  instead, the target state and covariance are estimated simultaneously online.
  \hfill $\blacksquare$
\end{remark}

Denote by~$n\in \overline{\mathcal{N}}\subset \mathcal{N}$ if robot~$n$ becomes active.
There are two constraints for its choice of actions~$\mathbf{s}_n(t)$, i.e.,
\begin{equation}\label{eq:sense-limit}
d(\mathbf{x}_{n}(t), \, \mathbf{y}_{m}(t))<R_n,\;   |\mathbf{s}_n(t)|<C_n;
\end{equation}
where (i)~$d(\cdot)$ is a distance measure and~$R_n>0$, as
typical sensors such as radar and {lidar}, are only valid within a limited range.
Outside this range, its uncertainty~$V(t)$ in~\eqref{eq:observation-model} is set to infinity.
Note that $R_n$ might be different across the robots;
(ii) there is often \emph{maximum} number of targets each robot~$n$
can track simultaneously due to limited computation or energy resources,
denoted by~$C_n\geq 1$.

\begin{remark}\label{remark:active}
  Note that the definition of action~$\mathbf{s}_n(\cdot)$ is essential,
  and different from most existing work.
  Namely, some work~\cite{tzes2023graph,atanasov2014information,schlotfeldt2018anytime}
  and our previous work~\cite{kantaros2019asymptotically,kantaros2021scalable}
  assume that the robot can obtain measurements of \emph{all} targets
  at all time,
  while others~\cite{sung2020distributed, tokekar2014multi, zhou2019sensor} limit
  to only \emph{one} target.
  In practice, there is often either {an} energy or computation cost associated with detection and tracking one target.
  Thus, the robot has to actively decide which targets to measure within its sensing range,
  under a maximum number.
  \hfill $\blacksquare$
\end{remark}

\subsection{Target State Estimation and Tracking}\label{subsec:estimate-track}
Initially, each target~$m \in \mathcal{M}$ has a prior Gaussian distribution for its
initial state $\mathbf{y}_m(0) \sim \mathcal{G}(\mathbf{\mu}^0_m,\,\Sigma^0_m)$,
which is known to all robots.
For time~$t>0$,
each robot~$n\in \mathcal{N}$ can move around
and make observations of some targets via~\eqref{eq:observation-model}, i.e.,
$\mathbf{Z}_m(t) \triangleq \{\mathbf{z}_{nm}(t),\, \forall n\in \mathcal{N}\}$,
which may contain time-stamped observations of all targets if measured,
otherwise set to empty.
Given the accumulated measurements~$\mathbf{Z}_m(0:t)$,
the posteriori distribution of target~$m$ is denoted by:
\begin{equation}\label{eq:target-posteriori}
  \widehat{\mathbf{y}}_m(t) \sim \mathcal{G}\big(\mathbf{\mu}_m(t),\,\Sigma_m(t)\big),
\end{equation}
which is the estimated mean and covariance of target~$m$ at time~$t$,
e.g., conveniently by the
EKF~\cite{atanasov2014information,athans1972determination}
when the target moves along one fixed road as described in the sequel.
However, if target~$m$ is \emph{not} monitored when it crosses a $K$-way intersection,
its estimation~$\widehat{\mathbf{y}}_m(t)$ changes by approximating
the $K$ possible directions statistically, i.e.,
\begin{equation}\label{eq:k-way}
  \mathbf{\mu}_m(t),\, \Sigma_m(t)  \triangleq \mathcal{G}_{\texttt{apx}}\Big(\big\{\mathbf{\mu}^k_m(t),
  \,k=1,\cdots,K\big\}\Big),
\end{equation}
where~$\mathcal{G}_{\texttt{apx}}(\cdot)$ is the Gaussian approximation of a set of points;
$\|\widehat{\mu}_{m}(t-1)-\nu\|<\varepsilon$, for some intersection~$\nu\in \mathcal{V}$
and a small margin~$\varepsilon>0$;
and $\mathbf{\mu}^k_m(t)$ is the predicted next state if target~$m$ follows the $k$-th direction at time~$t-1$.
In other words, the uncertainty increases drastically if a target crosses intersections
without being monitored.
Thus, the determinant of the covariance matrix $\textbf{det}\big(\Sigma_m(t)\big)$ is
used to evaluate the quality of state estimation for target~$m$ up to time~$t$.

\subsection{Problem Statement}\label{subsec:prob-statement}
The considered problem can be stated as a long-term constrained optimization problem, i.e.,
\begin{subequations}\label{eq:problem}
\begin{align}
  & \min_{\overline{\mathcal{N}}(t),\, \{(\mathbf{u}_n, \mathbf{s}_n)\}}\, \left\{ \lim_{T\rightarrow \infty} \frac{1}{T} \, \sum^T_{t=0}\, \overline{N}(t) \right\} \notag\\
  \textbf{s.t.}\quad & \textbf{det}\big(\Sigma_m(t)\big) < \gamma_m, \;\forall m, \,\forall t;\label{subeq:bound}\\
  &\eqref{eq:robot-model}-\eqref{eq:k-way},\;\forall n,\,\forall m,\, \forall t; \label{subeq:model}
\end{align}
\end{subequations}
where~$\overline{N}(t)>0$ is the number of active robots in~$\overline{\mathcal{N}}(t)$ for~$t\geq 0$;
the objective is the average number of active robots in the long horizon;
$\gamma_m>0$ is a user-specified bound on the estimation uncertainty for each target~$m \in \mathcal{M}$ in~\eqref{subeq:bound};
constraints in~\eqref{subeq:model} are the aforementioned dynamic and observation models, and how the estimation uncertainties of the targets evolve under the measurements.

Note that $\gamma_m$ above is a key design parameter that greatly influence the system behavior.
For instance, if $\gamma_m$ is the minimum variance that can be achieved if $N_0$ robots
monitor one target at all time, then the required team size is at least $N_0M/(\max_{m}\{C_m\})$.
On the other hand, if $\gamma_m$ is sufficiently large, one robot could suffice
by observing each target in sequence.
More numerical analyses are given in Sec.~\ref{sec:experiments}.

\begin{remark}\label{rm:objective}
  The objective in~\eqref{eq:problem} boils down to extending the longevity of each robot,
  e.g., when a UAV is inactive for the monitoring task,
  it can hover, glide or even park to charge battery
  until being active next time.
 \hfill $\blacksquare$
\end{remark}

\section{Proposed Solution}\label{sec:solution}
Due to the dynamic nature of the proposed problem,
the proposed solution consists of two hierarchical components,
i.e., the dynamic assignment of monitoring tasks
based on the estimated target behaviors with a minimum fleet size;
and the optimal control of each robot given the assigned targets
to ensure the estimation uncertainty is bounded.
Both components are executed in a receding horizon fashion,
but with different frequencies and horizons.

\subsection{Dynamic Assignment of Monitoring Tasks}\label{subsec:assignment}

\subsubsection{Prior and Posteriori Estimation of Target Trajectory}\label{subsubsec:EKF}
Let the estimated state of target~$m$ be~$(\mu_m(t_0),\,\Sigma_m(t_0))$ at time~$t_0\geq 0$.
Then, assuming that \emph{no} measurements are obtained during the future
period~$t\in [t_0,\,t_0+T_{\texttt{s}}]$ for $T_{\texttt{s}}>0$,
the predicted evolution of mean and covariance is given by:
\begin{equation}\label{eq:prior}
  \begin{split}
    &\widehat{\mu}_{m}(t+1) = G\, \widehat{\mu}_{m}(t),\\
    &\widehat{\Sigma}_{m}(t+1) = G\, \widehat{\Sigma}_{m}(t)\, G + W_m,\\
    &(\widehat{\mu}_{m}(t+1),\, \widehat{\Sigma}_{m}(t+1)) =
      \mathcal{G}_{\texttt{apx}}\big(\{\widehat{\mu}^k_{m}(t)\}\big),\\
\end{split}
\end{equation}
where~$(\widehat{\mu}_{m},\, \widehat{\Sigma}_{m})$ are the predicted mean and covariance;
$W_m$ is the uncertainty from~\eqref{eq:target-model};
the last case is when target~$m$ is predicted to cross an intersection from~\eqref{eq:k-way}.
Thus, the prior estimation of target~$m$ during~$[t_0,\,t_0+T_{\texttt{s}}]$ is
denoted by~$\widehat{\boldsymbol{\mu}}_m(t)$ and $\widehat{\boldsymbol{\Sigma}}_m(t)$; and $\widehat{\mathbf{y}}_m(t)\triangleq(\widehat{\boldsymbol{\mu}}_m(t),\,\widehat{\boldsymbol{\Sigma}}_m(t)$.

\begin{figure}[t]
  \centering
  \includegraphics[width=0.85\linewidth]{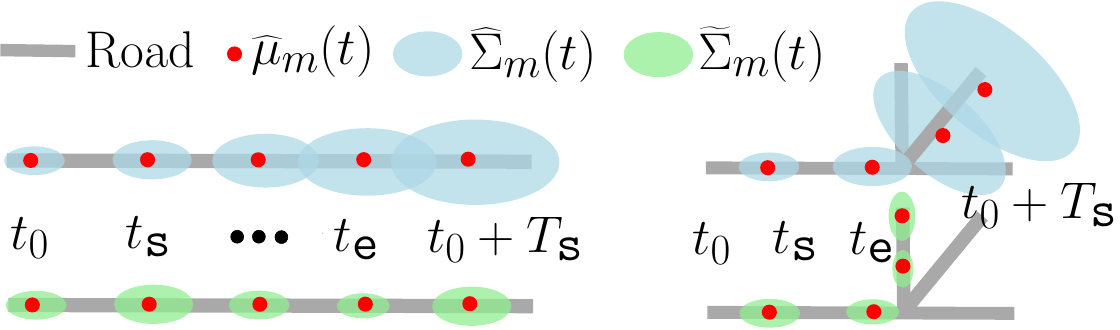}
  \vspace{-3mm}
  \caption{Illustration of how the posteriori uncertainty~$\widetilde{\Sigma}_m$ changes
    by~\eqref{eq:total-post}, compared with the prior~$\widehat{\Sigma}_m$:
    on one road (\textbf{Left}) and an interaction (\textbf{Right}).}
  \label{fig:widetilde-sigma}
  \vspace{-6mm}
\end{figure}

Then, assume that one robot~$n$ actively monitors target~$m$ during the
period~$[t_{\texttt{s}},t_{\texttt{e}}]$,
where~$t_0+T_{\texttt{s}}>t_{\texttt{e}}>t_{\texttt{s}}>t_0$.
The resulting change in its covariance is given by the
EKF~\cite{atanasov2014information,athans1972determination,barfoot2017state} via the prediction and update as follows:
\begin{equation}\label{eq:post}
  \begin{split}
    &{\Sigma}'_{m}(t+1) = G\, \widetilde{\Sigma}_{m}(t)\, G + W_m,\quad
      \forall t\in [t_{\texttt{s}},\, t_{\texttt{e}}]; \\
      &\widetilde{\Sigma}_{m}(t+1) = {\Sigma}'_{m}(t) - K_m(t) H_m(t) {\Sigma}'_{m}(t);\\
      & K_m(t) = {\Sigma}'_{m}(t)
    H_m^{\intercal}(t) R_m^{-1}(t)\\
    & R_m(t) = H_m(t) {\Sigma}'_{m}(t)H_m^{\intercal}(t)  + V(\mathbf{x}_n(t),\,
    \widehat{\mu}_{m}(t)); \\
    & H_m(t) = \nabla_{\mathbf{y}}h(\mathbf{x},\,\mathbf{y})|_{(\mathbf{x}_n(t),\, \widehat{\mu}_{m}(t))};
\end{split}
\end{equation}
where~$\mathbf{x}_n(t)$ is the robot state;
$\widetilde{\Sigma}_{m}(t)$ is the posteriori estimation uncertainty
assuming that robot~$n$ starts monitoring target~$m$ from time~$t_{\texttt{s}}$;
$H_m(t)$ is the linearized observation model of~$h(\cdot)$ given~$\mathbf{x}_n(t)$
and~$\widehat{\mu}_{m}(t)$;
$V(\cdot)$ is the state-dependent uncertainty during observation
from~\eqref{eq:observation-model};
and~$\widetilde{\Sigma}_m(t_{\texttt{s}})$ is initialized according to the prior estimation.
Afterwards, for~$t>t_{\texttt{e}}$ without measurements,
$\widetilde{\Sigma}_m(t)$ evolves in the same way as~\eqref{eq:prior}.
In other words, the posteriori estimation $\widetilde{\Sigma}_m(t)$ is
computed in three segments as follows:
\begin{equation}\label{eq:total-post}
  \begin{split}
    &\widetilde{\Sigma}_m(t) = \widehat{\Sigma}_m(t), \; \forall t\in [t_0,\,t_{\texttt{s}});\\
    & \widetilde{\Sigma}_m(t+1) = \rho_{\texttt{post}}(\widetilde{\Sigma}_m(t),\mathbf{x}_n(t)), \; \forall t\in [t_{\texttt{s}},\,t_{\texttt{e}});\\
    & \widetilde{\Sigma}_m(t+1) = \rho_{\texttt{pri}}(\widetilde{\Sigma}_m(t)), \; \forall t\in [t_{\texttt{s}},\, t_0+T_{\texttt{s}}];\\
\end{split}
\end{equation}
where functions~$\rho_{\texttt{post}}(\cdot)$ and $\rho_{\texttt{pri}}(\cdot)$
encapsulate~\eqref{eq:post} and~\eqref{eq:prior}, respectively.
An illustration is given in Fig.~\ref{fig:widetilde-sigma}.

\subsubsection{Formulation of Task Assignment Problem}\label{subsubsec:task-form-prob}
Clearly, the computation of~$\widetilde{\Sigma}_{m}(t)$ above depends on the robot
trajectory~$\mathbf{x}_n(t)$.
Due to the inherent combinatorial complexity of the assignment process,
an approximation method is introduced here to bound the duration required
for robot~$n$ to monitor target~$m$ without specifying the trajectory~$\mathbf{x}_n(t)$,
such that the uncertainty measure~$\textbf{det}(\widetilde{\Sigma}_{m}(t))$
above is reduced below~$\gamma_m$ in~\eqref{subeq:bound}.
Instead, the exact optimization of robot trajectories is performed in
Sec.~\ref{subsec:nmpc} afterwards.
More specifically, consider first the following assumption:
\begin{assumption}\label{eq:single}
  For arbitrary~$(\widehat{\boldsymbol{\mu}}_m(t),\, \widehat{\boldsymbol{\Sigma}}_m(t))$
  above such that $\widehat{\boldsymbol{\Sigma}}_m(t_0)<\gamma_m$ and a large enough~$T_{\texttt{s}}$,
  there always exists a period~$[t_{\texttt{s}},\, t_{\texttt{e}}]$
  and a choice of~$\mathbf{x}_n(t)$
  such that $\textbf{det}(\widetilde{\Sigma}_m(t))<\gamma_m$, $\forall t\in [t_{\texttt{e}},\,t_0+T_{\texttt{s}}]$.
\hfill $\blacksquare$
\end{assumption}
The above assumption basically states that \emph{one robot is \textbf{sufficient} to monitor one target
  while ensuring that its uncertainty stays below the given bound}.
This assumption stems from the application that a few UAVs {are} deployed to monitor
a large number of dynamic targets.
If more than one {robot is} required to monitor even a single target,
then there are more robots than targets, which is not the focus here.

More specifically,
the starting time~$t_{\texttt{s}}$ is chosen
between~$[t_0,\,\overline{t}_{\texttt{s}}]$, where~$\overline{t}_{\texttt{s}}$ is
the \emph{latest} starting time such that:
$\widehat{\Sigma}_m(\overline{t}_{\texttt{s}}-1)<\gamma_m$
and
$\widehat{\Sigma}_m(\overline{t}_{\texttt{s}})\geq \gamma_m$
hold.
Given a choice of~$t_{\texttt{s}}$, the ending time~$t_{\texttt{e}}$ is minimized
to monitor more targets, i.e., as the \emph{earliest} instant such that:
$\mathbf{x}_n(t) = \min_{\mathbf{x}_n(t)}\, \textbf{det}(\widetilde{\Sigma}_m(t+1)),
\;\forall t\in [t_\texttt{s},\, t_\texttt{e}]$
and
$\textbf{det}(\widetilde{\Sigma}_m(t_\texttt{s}+T_{\texttt{s}})) <\gamma_m$ hold,
where $\widetilde{\Sigma}_m(t)$ is determined by~\eqref{eq:total-post};
the first condition enforces that robot~$n$ chooses the best observation position
to reduce the uncertainty the most during each step of the monitoring
period~$(t_{\texttt{s}},\, t_{\texttt{e}})$;
the second condition requires that the terminal uncertainty is below~$\gamma_m$.
If both conditions hold, the choice of schedule~$(t_{\texttt{s}},\, t_{\texttt{e}})$
is called feasible.
Thus, for a robot-target pair $(n,\, m)$,
the time to start monitoring~$t_{\texttt{s}}$ has a window~$[t_0,\,\overline{t}_{\texttt{s}}]$.
The monitoring plan for each robot~$n$ is defined as:
\begin{equation}\label{eq:schedule}
  \boldsymbol{\tau}_n\triangleq
  (t^{m_1}_{\texttt{s}},\, t^{m_1}_{\texttt{e}}) \cdots (t^{m_L}_{\texttt{s}},\, t^{m_L}_{\texttt{e}}),
\end{equation}
where~$(t^{m_\ell}_{\texttt{s}},\,t^{m_\ell}_{\texttt{e}})$  is the schedule
for monitoring target~$m_\ell\in \mathcal{M}$, $\forall \ell =1,\cdots,L$.
The plan length is determined by the elapsed time~$t^{m_L}_{\texttt{e}}-t^{m_1}_{\texttt{s}}$.
A complete plan is \emph{feasible} if all contained schedules are feasible.

\begin{problem}\label{prob:assignment}
  Given the prediction of all targets $\big\{(\widehat{\boldsymbol{\mu}}_m(t),\, \widehat{\boldsymbol{\Sigma}}_m(t)),\forall m \big\}$,
  determine first (I) the minimum number of active robots~$\overline{N}(t)$;
  and then (II) a feasible monitoring plan~$\boldsymbol{\tau}_n$
  for each~$n\in \overline{\mathcal{N}}(t)$,
  such that the maximum plan length of all robots is minimized.
  \hfill $\blacksquare$
\end{problem}

\subsubsection{Solution Based on m-MVRPTW}\label{subsubsec:m-mvrptw}

\begin{figure}[t]
  \centering
  \includegraphics[width=0.6\linewidth]{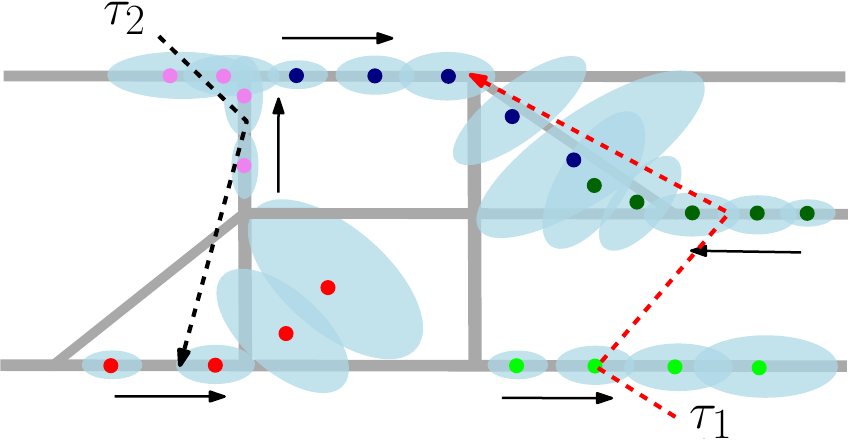}
  \vspace{-3mm}
  \caption{Illustration of the assignment algorithm given the predicted trajectories with covariances,
  i.e., $5$ targets monitored by $2$ robots.}
  \label{fig:assign}
  \vspace{-5mm}
\end{figure}

The solution to Problem~\ref{prob:assignment} is based on formulating it as a flexible-size
multiple vehicle routing problem with time windows
(m-MVRPTW)~\cite{polacek2004variable,repoussis2010solving}.
In particular, three steps are followed:
(I) A service location is created for each target~$m\in \mathcal{M}$
at~$p_m\triangleq \widehat{\mu}_m(\hat{t}_m)$ for~$\hat{t}_m=(t^m_0+\overline{t}^m_{\texttt{s}})/2$,
with a \emph{strict} time window~$w_m\triangleq [t^m_0,\, \overline{t}^m_{\texttt{s}}]$.
Its service time~$T_m$ is given by:
  $T_m \triangleq \max_{t'_{\texttt{s}}\in [t^m_0,\, \overline{t}^m_{\texttt{s}}]}\,
  |t'_{\texttt{e}}-t'_{\texttt{s}}|$,
where~$t'_{\texttt{s}}$ is a valid starting time and $t'_{\texttt{e}}$ is the associated ending time;
$T_m$ is the longest duration for target~$m$;
(II) The transition time from target~$m_1$ to target~$m_2$ is computed as:
  $T_{m_1m_2} \triangleq \max_{t^{m_1}_{\texttt{e}},t^{m_2}_{\texttt{s}}}
  \, \{\|\widehat{\mu}_{m_1}(t^{m_1}_{\texttt{e}})-\widehat{\mu}_{m_2}(t^{m_2}_{\texttt{s}})\|/v_n\}$,
where~$t^{m_1}_{\texttt{e}}$ is the allowed ending time for target~$m_1$;
$t^{m_2}_{\texttt{e}}$ is the allowed starting time for target~$m_2$;
and~$v_n$ is the average velocity of robot~$n$;
(III) The m-MVRPTW problem is formulated as
$M$ service locations at~$\{p_m,\forall m\}$ with a strict time window~$\{w_m,\forall m\}$,
among which the time matrix is given by~$\{T_{m_1}+T_{m_1m_2},\forall m_1,m_2\}$.
Each robot can depart from its current location and end anywhere.
The proposed algorithm gradually adds robots to the active fleet,
ranked by their minimum distance to any target.
Given a fixed fleet size, the standard MVRPTW can be solved
by any solver, e.g., \texttt{OR-Tools}~\cite{gor}.
The outputs are the set of required
robots~$\overline{\mathcal{N}} \subseteq \mathcal{N}$ with {cardinality}~$\overline{N}$;
and their schedule~$\boldsymbol{\tau}_n$ from~\eqref{eq:schedule},
with the sequence of assigned targets $\mathbf{m}_n\triangleq m_1\cdots m_L$
and the associated service schedule $\mathbf{t}_n$, $\forall n\in \overline{\mathcal{N}}$.




\subsection{Informative Trajectory Planning via Distributed NMPC}\label{subsec:nmpc}

Given the schedule~$\{\boldsymbol{\tau}_n,\forall n\in \overline{\mathcal{N}}\}$,
each robot~$n$ should optimize its actual trajectory to execute this schedule,
while ensuring the uncertainty constraints.
The main challenges are two-fold:
(i) both the robot dynamics and the uncertainty bounds are nonlinear
and (ii) the target behavior is unknown and only observed online.
Thus, the framework of distributed nonlinear model predictive control (NMPC)
is applied~\cite{allgower2012nonlinear,kang2009linear},
for the trajectory optimization problem below.

\begin{problem}\label{prob:nmpc}
  Given the schedule~$\boldsymbol{\tau}_n$,
  compute the optimal control and action~$(\mathbf{u}_n,\,\mathbf{s}_n)$
  for each active robot~$n\in \overline{\mathcal{N}}$,
  such that the uncertainties of all assigned targets~$\mathbf{m}_n$ are bounded
  and the control cost is minimized.
  \hfill $\blacksquare$
\end{problem}

Consider another planning horizon~$T_{\texttt{c}}>0$,
which is different and shorter than the horizon~$T_{\texttt{s}}$
of task assignment in Sec.~\ref{subsec:assignment}.
Given the prediction of target behaviors~$\big\{(\widehat{\boldsymbol{\mu}}_m
(t_0),\,\widehat{\boldsymbol{\Sigma}}_m(t_0)),\forall m \in \mathbf{m}_n\big\}$
at time~$t_0$, the NMPC problem is formulated for each robot~$n\in \overline{\mathcal{N}}$
as follows:
\begin{equation}\label{eq:nmpc}
  \begin{split}
    &\min_{(\mathbf{u}_n,\,\mathbf{s}_n)}\; \sum_{t\in \mathcal{T}_{\texttt{c}}}
    \Upsilon(\mathbf{x}_n(t),\, \mathbf{u}_n(t))\\
    \textbf{s.t.} \quad & \mathbf{x}_{n}(t+1) \triangleq f\big(\mathbf{x}_{n}(t),
    \, \mathbf{u}_{n}(t)\big);\\
    & |\mathbf{s}_n(t)| < C_n;\;
    \textbf{det}\big(\widetilde{\Sigma}_m(t)\big) < \gamma_m;\\
    & \widetilde{\Sigma}_m(t+1) = \rho\big(\widetilde{\Sigma}_m(t),\,
    \mathbf{x}_n(t),\, \widehat{\mathbf{y}}_m(t_0)\big);
    \end{split}
\end{equation}
where $\mathcal{T}_{\texttt{c}}\triangleq \{t_0,\cdots,t_0+T_{\texttt{c}}\}$;
$\Upsilon: \mathcal{W} \times \mathcal{U} \rightarrow \mathbb{R}^+$
is a general function to measure the control cost given a choice of
control input at certain states;
the uncertainty measure~$\widetilde{\Sigma}_m(t)$ is the posteriori
estimation from~\eqref{eq:post};
function~$\rho$ summarizes the updating rules from~\eqref{eq:total-post}
given the current prediction~$\widehat{\mathbf{y}}_m(t_0)$;
and these constraints hold for all $t\in \mathcal{T}_{\texttt{c}}$
and $m\in \mathbf{m}_n$.

Note that the above constrained optimization is different from the
standard NMPC in~\cite{allgower2012nonlinear} due to the variable~$\mathbf{s}_n$
as the monitoring action to decide which targets to monitor,
which is discrete thus yielding a combinatorial complexity over~$\mathbf{m}_n$.
Instead, a hierarchical structure is followed to decouple the optimization
over~$\mathbf{s}_n$ and~$\mathbf{u}_n$.
Specifically, given the current states of robot~$n$ and all targets
in~$\mathbf{m}_n$ at each time step~$t_0$,
the \emph{first}~$C_n$ targets in~$\mathbf{m}_n$ with the largest
predicted prior uncertainty~$\widehat{\Sigma}_m$
within~$\mathcal{T}_{\texttt{c}}$, i.e.,
\begin{equation}\label{eq:best-k}
  \mathbf{s}_n(t) = \underset{k=C_n,m\in \mathbf{m}_n}{\textbf{max\_k}}
  \Big \{\max_{t\in \mathcal{T}_c}\, \textbf{det}(\widehat{\Sigma}_m(t))\Big\},
  \;\forall t \in \mathcal{T}_c;
\end{equation}
where~$\mathbf{max\_k}(\cdot)$ is the operator for choosing the~$k$
largest elements in a set;
$\mathbf{s}_n(t)$ is kept constant in the optimization~\eqref{eq:nmpc}.
Note that if~$|\mathbf{m}_n|\leq C_n$, all targets are monitored.
Once~$\mathbf{s}_n$ is determined,
the optimization~\eqref{eq:nmpc} is transformed into a nonlinear
optimization problem, where the variables are:
(i) the robot state~$\mathbf{x}_n(t)$ and control input~$\mathbf{u}_n(t)$
at each step~$t \in \mathcal{T}_c$;
and (ii) the posteriori uncertainty~$\widetilde{\Sigma}_m(t)$ at each step
$t \in \mathcal{T}_c$ and for each target~$m\in \mathbf{s}_n$.
The determinant $\textbf{det}(\widetilde{\Sigma}_m(t))$ of a square matrix
can be computed \emph{analytically}~\cite{horn2012matrix}, i.e.,
as a polynomial in the traces of the powers of $\widetilde{\Sigma}_m(t)$.
Thus, the formulated problem can be solved by off-the-shelf solvers,
e.g., \texttt{CasADi}~\cite{Andersson2019}.
The results are the desired control~$\mathbf{u}_n(t)$ in the planning horizon.

\begin{remark}\label{rm:nmpc}
  Related work often assumes
  either a discretized control space~\cite{tzes2023graph, atanasov2014information, kantaros2019asymptotically, guo2016task}
  or pre-defined motion primitives~\cite{tokekar2014multi, zhou2019sensor, zhou2023robust},
  which {cannot} be applied here for trajectory optimization under dynamic constraints.
 \hfill $\blacksquare$
\end{remark}


\setlength{\textfloatsep}{5pt}
\begin{algorithm}[t]
  \caption{Uncertainty-bounded Monitoring} \label{alg:overall}
  \LinesNumbered
  \KwIn{$\mathcal{M}$, $\mathcal{R}$, $\{\widehat{\mathbf{y}}_m(0)\}$,
    $T_{\texttt{s}}$, $T_{\texttt{c}}$.}
  \KwOut{$\overline{\mathcal{N}}(t)$, $\{\widehat{\mathbf{y}}_m(t)\}$, $\{(\mathbf{u}_n,\,\mathbf{s}_n)\}$.}
  \While{not terminated}{
    Compute~$\{\widehat{\mathbf{y}}_m(t)\}$ by~\eqref{eq:prior}\;
    Formulate m-MVRPTW by Sec.~\ref{subsubsec:m-mvrptw}\;
    Derive~$\overline{\mathcal{N}}(t)$ and $\{(\mathbf{m}_n,\mathbf{t}_n)\}$, e.g., via~\cite{gor}\;
    Formulate NMPC~\eqref{eq:nmpc} for each robot~$n\in \overline{\mathcal{N}}(t)$\;
    Derive solution~$\{(\mathbf{u}_n,\,\mathbf{s}_n)\}$, e.g., via~\cite{Andersson2019}\;
    Each robot~$n\in \overline{\mathcal{N}}(t)$ applies~$(\mathbf{u}_n(t),\mathbf{s}_n(t))$\;
    Make observations~$\{\mathbf{z}_{nm}\}$\;
    Update~$\{\widehat{y}_m(t+1)\}$ by~\eqref{eq:post-update}\;
    $t\leftarrow t+1$\;
  }
\end{algorithm}

\subsection{Overall Framework}\label{subsec:overall}
As summarized in Alg.~\ref{alg:overall},
the above two modules are executed online in a receding horizon
fashion with different frequencies.
More specifically,
at each time step~$t>0$,
the prior estimation~$\widehat{\mathbf{y}}_m(t)$ is computed
via~\eqref{eq:prior} for each target. Based on this estimation,
the monitoring tasks are assigned to a subset of robots by
solving~\eqref{eq:prior}, yielding the planned
schedule~$\boldsymbol{\tau}_n$.
Then, the optimization for control and monitoring action
is formulated via~\eqref{eq:nmpc}
given~$\boldsymbol{\tau}_n$ and $\widehat{\mathbf{y}}_m(t)$,
yielding the optimal control sequence~$\mathbf{u}_n$
and the monitored action~$\mathbf{s}_n$.
Afterwards, each robot moves by executing the next control
input~$\mathbf{u}_n(t)$,
while obtaining possible new measurements~$\{\mathbf{z}_{nm},\forall m\in \mathbf{s}_n\}$.
Given these measurements, the posteriori
estimation~$\widehat{y}_m(t+1)=(\widehat{\mu}_m(t+1),\widehat{\Sigma}_m(t+1))$
is updated for each observed target as follows:
\begin{equation}\label{eq:post-update}
  \begin{split}
  \widehat{\mu}_m(t+1) &= \widehat{\mu}_m(t) + K_m\big{(}\mathbf{z}_{nm}
    -h(\mathbf{x}_n(t), \widehat{\mu}_m(t))\big{)},\\
    \widehat{\Sigma}_m(t+1) &= \rho_{\texttt{post}}(\widehat{\Sigma}_m(t),\, \mathbf{x}_n(t)),
\end{split}
\end{equation}
where~$K_m$ is computed as in~\eqref{eq:post}
and~$\rho_{\texttt{post}}(\cdot)$ defined in~\eqref{eq:total-post}.
This procedure is repeated until the system is terminated.
It is worth noting that while the trajectory optimization is
executed at every time step locally by each robot,
the module of task assignment is executed at a much lower frequency
via a central coordinator,
which can be periodic or event-based, e.g.,
each time the set~$\mathbf{s}_n$ computed in~\eqref{eq:best-k} changes.

Lastly, note that even as approximated sub-problems, both Problems~\ref{prob:assignment}
and~\ref{prob:nmpc} have high computation complexity.
In particular, Problem~\ref{prob:assignment} remains NP-hard~\cite{polacek2004variable}
as being a more general problem than the classic MVRP,
while there is no guarantee that Problem~\ref{prob:nmpc} converges to the global
optimal solution~\cite{wright2006numerical} as being nonlinear and strongly non-convex.
Nonetheless, for moderate problem sizes e.g.,~$10$ robots and~$10$ targets,
these problems can be solved in $0.89s$ and $0.54s$.
More details are given in the next section.


\section{Numerical Experiments} \label{sec:experiments}

Extensive large-scale simulations are presented in this section.
The proposed method is implemented in Python3 and tested on a laptop with an Intel Core i7-1280P CPU.
Simulation videos can be found in the supplementary files.

\begin{figure}[t]
  \centering
  \includegraphics[width=0.95\linewidth]{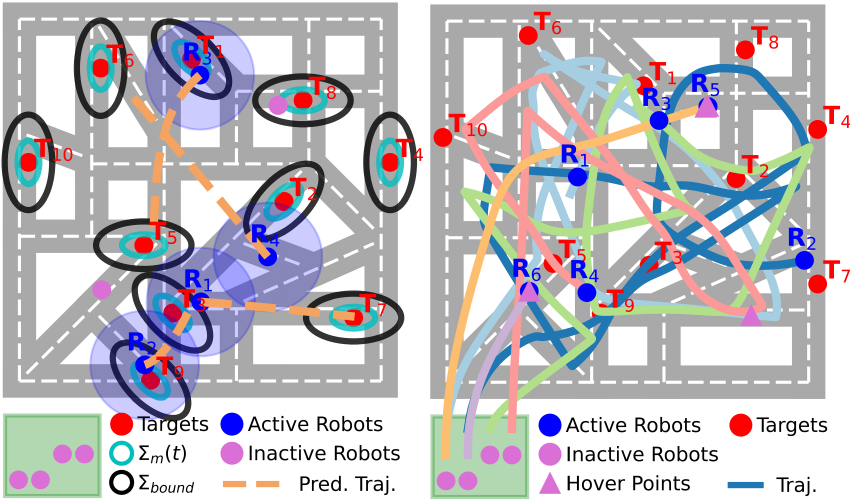}
  \vspace{-2mm}
  \caption{\textbf{Left}: Snapshot at $t=52s$ where $4$ robots are active.
    \textbf{Right}: Final trajectories of all robots with marked status.}
  \label{fig:snap}
  \vspace{-1mm}
\end{figure}

\subsection{Description}\label{subsec:description}
As shown in Fig.~\ref{fig:snap},
consider~$10$ targets moving on the road-networks with a size of~$10m \times 10m$,
and there are~$10$ robots in the fleet.
Each robot follows the standard unicycle dynamics for UAV flying at a fixed height.
The observation model is the commonly-used range-and-bearing
model~\cite{schlotfeldt2018anytime,kantaros2019asymptotically},
where the uncertainty $V$ in~\eqref{eq:observation-model} grows linearly in range and bearing.
The exact mathematical model is omitted here due to limited space.
The limited range in~\eqref{eq:sense-limit} is set to $R_n=1.5m$
and the capacity $C_n=5$ uniformly for all robots.
Initially, the target states are chosen \emph{randomly} along the roads with a small uncertainty,
while the robots start from the base.
During execution, the targets choose the next road randomly at intersections,
which are unknown to the robots.
The threshold for uncertainty is set to $\gamma_m = 0.1$ for all targets.
Moreover, the horizon for dynamic assignment is set to~$T_{\texttt{s}} = 50$
and updated every $10s$,
while the NMPC planning horizon is set to~$T_{\texttt{c}} = 10$ and updated every $0.1s$.
The simulation is terminated at $100s$.
\subsection{Results}\label{subsec:results}
The final results are shown in Fig.~\ref{fig:overall} and~\ref{fig:snap}.
The assignment method via $m$-MVRPTW takes in average $0.85s$,
which generates a varying number of active robots (minimum $3$ and maximum $6$).
For instance, it is enough at $31s$ for $3$ robots to monitor all $10$ targets,
while the rest becomes inactive.
At $50s$, more robots are recruited given the predicted uncertainty.
Moreover, the average computation time of NMPC is $0.57s$
and the resulting trajectory often covers multiple targets simultaneously.
The estimation {uncertainties} of all targets are below the specified bound~$0.1$,
while the average number of active robots is $4.3$ and the average number of targets assigned to
each robot is $2.3$.
The final trajectories for all the robots are depicted in Fig.~\ref{fig:snap},
which switches between being active and inactive.

\subsection{Comparisons}\label{subsec:comparisons}

\begin{figure}[t]
  \centering
  \includegraphics[width=0.98\linewidth]{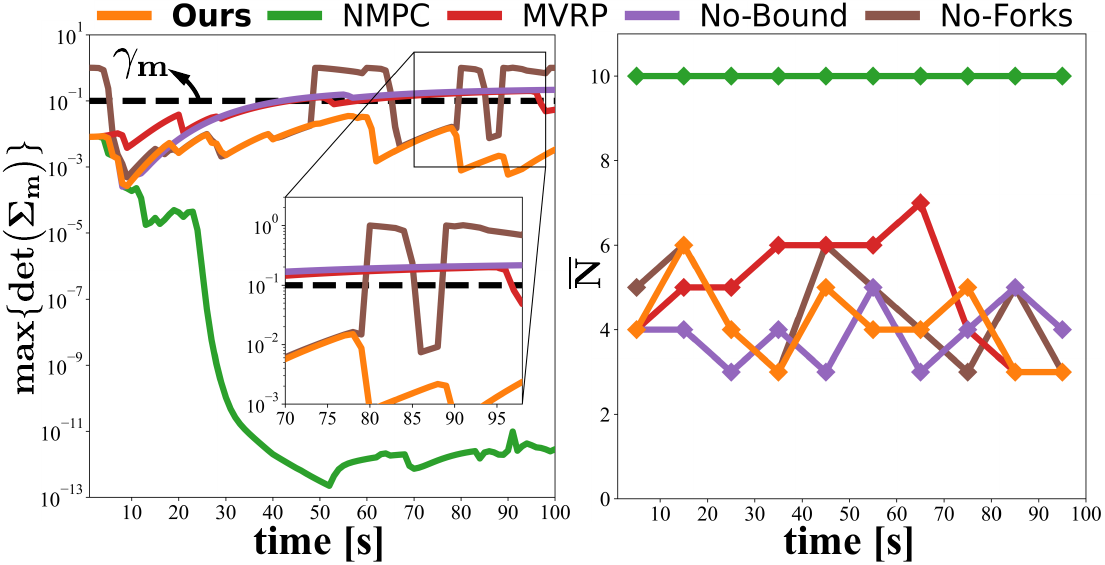}
  \vspace{-2mm}
  \caption{
    Comparison against four baselines:
    the maximum uncertainty of all targets (\textbf{Left})
    and the number of active robots within the fleet (\textbf{Right}).}
  \label{fig:compare}
  \vspace{-1mm}
\end{figure}

To further validate the effectiveness of the proposed framework (as \textbf{Ours}),
a quantitative comparison is conducted against four baselines
(two ablation studies and two common approaches):
(i) \textbf{NMPC}, where the first layer of task assignment is skipped
and each robot is assumed to track one single target at each time step;
(ii) \textbf{MVRP}, where the proposed assignment algorithm remains
but the trajectory optimization is replaced by a simple strategy that
tracks the assigned targets in order;
(iii) \textbf{No-Bound}, where the objective in~\eqref{eq:nmpc}
is changed to minimize the summed uncertainties of all targets
(see~\cite{atanasov2014information,dames2012decentralized,schlotfeldt2018anytime}),
i.e., not as a hard constraint;
(iii) \textbf{No-Forks}, where the target model follows~\eqref{eq:target-model}
but without considering the unknown behavior at intersections.
The compared metrics are the planning time, the success rate, the maximum uncertainty of all targets,
and finally the average number of active robots.

The results are summarized in Fig.~\ref{fig:compare}.
It can be seen that
both the proposed method and the {NMPC} can ensure that the estimation
uncertainty is below the given bound at all time, while others fail.
However, the {NMPC} requires that all robots are active at all time,
while ours requires in average $4$ robots.
In contrast, the {MVRP} method optimizes the task assignment,
but the estimation uncertainty might exceed the bound
as the motion strategy {cannot} react to all assigned targets.
Similarly, despite of faster planning,
the {No-Bound} method loses the guarantee on uncertainty as the hard constraints are relaxed.
At last, the necessity of modeling intersection is clear with the {No-Forks} method,
where the uncertainty increases drastically above the bound when the targets cross intersections.
Our method however can predict this change and monitor these targets in advance.


\subsection{Scalability Analysis}\label{subsec:scalability}
Scalability of the proposed algorithm is analyzed w.r.t. three aspects:
the number of targets $M$, the upper bound of uncertainty~$\gamma_m$, and the capacity~$C_n$.
The recorded metrics are similar to the previous part, including the detailed planning 
time for MVRP and NMPC.
The results are summarized in Table~\ref{table:table-data},
where the nominal setup is from Sec.~\ref{subsec:results}.
Notably, when $M$ is increased to $20$ and $100$,
the success rate remains $100\%$.
Planning time of MVRP is increased significant as the number of tasks has increased to $100$,
while the NMPC takes roughly the same time as each robot solve it locally.
It is worth noting that the ratio between
the average number of active robots and the total number of targets remains constant,
meaning that the \emph{fleet efficiency} does not degrade as $M$ increases.
Furthermore,
when the threshold~$\gamma_m$ is decreased to $10^{-3}$ and even $10^{-5}$,
the success rate decreases as the NMPC is harder to solve with a tighter bound,
yielding a much higher number of active robots.
Lastly, when $C_n$ is decreased to $3$, the planning time of NMPC is decreased as the number of constraints in~\eqref{eq:nmpc} is reduced,
and the average number of active robots is increased
as one robot can monitor less targets simultaneously.
The opposite holds for $C_n=7$, where the average number of active robots
is as low as $3.3$ out of $10$ in total.

\begin{table}[t]
 \begin{center}
 \begin{threeparttable}
   \caption{SCALABILITY ANALYSIS RESULTS}\label{table:table-data}
   \vspace{-0.05in}
   \setlength{\tabcolsep}{0.5\tabcolsep}
   \centering
   \begin{tabular}{cc|c c c c c}
     \toprule
     \textbf{Param.} & \textbf{Values} & \textbf{\makecell{Success \\ Rate [\%]}} & \textbf{\makecell{MVRP \\ Time [s]}} & \textbf{\makecell{NMPC \\ Time [s]}} & \textbf{\makecell{Average\\ $\overline{N}$}}\\
     \midrule
     \multicolumn{2}{c}{{Nominal}\tnote{*}} \vline & {100.0} & {0.80} & {0.57} & {5.0} \\
     \hline
     \multirow{2}{*}{$M$} & {20} & {100.0} & {1.37} & {0.54} & {10.3} \\
                          & {100} & {100.0} & {14.2} & {0.56} & {56.3} \\
     \hline
     \multirow{2}{*}{$\gamma_m$} & {$10^{-3}$} & {98.3} & {1.24} & {0.53} & {5.6} \\
                                 & {$10^{-5}$} & {83.8} & {4.43} & {0.67} & {7.7} \\
     \hline
     \multirow{2}{*}{$C_n$} & {3} & {100.0} & {0.74} & {0.40} & {6.6} \\
                            & {7} & {100.0} & {0.67} & {0.83} & {3.3} \\
     \bottomrule
   \end{tabular}
     \begin{tablenotes}[para,flushleft]
        \footnotesize
        \item[*] Nominal case: $M=10$, $\gamma_m=10^{-1}$, and $C_n=5$.
     \end{tablenotes}
 \end{threeparttable}
\end{center}
 \vspace{-3mm}
 \end{table}

\section{Conclusion} \label{sec:conclusion}
This work proposes an online task and motion coordination algorithm
for large-scale monitoring tasks over road-networks.
It ensures an explicitly-bounded estimation uncertainty for the targets,
while employing a minimum number of heterogeneous robots.
Future work involves a time-varying number of targets and complex tasks.


\newpage
\bibliographystyle{IEEEtran}
\bibliography{contents/references}

\end{document}